\documentclass[runningheads]{llncs}
\usepackage{graphicx}
\usepackage{multirow}

\usepackage{array}
\newcolumntype{P}[1]{>{\centering\arraybackslash}p{#1}}
\newcolumntype{C}[1]{>{\centering\arraybackslash}p{#1}}

\usepackage{graphicx}

\usepackage[normalem]{ulem}
\useunder{\uline}{\ul}{}

\begin{document}

\title{Don't lose the message while paraphrasing: \\ A study on content preserving style transfer}

\titlerunning{A study on content preserving style transfer}
%\orcidID{1111-2222-3333-4444} 

% \author{Anonymous submission}

\author{Nikolay Babakov\inst{1}\thanks{Work mostly has been done while at Skoltech} \and
David Dale\inst{3,\star} \and Ilya Gusev\inst{3}\and Irina Krotova\inst{4} \and 
\\Alexander Panchenko\inst{2,5}}
\authorrunning{N. Babakov et al.}

\institute{
Centro Singular de Investigación en Tecnoloxías Intelixentes (CiTIUS), Universidade de Santiago de Compostela \and
Skolkovo Institute of Science and Technology (Skoltech) \and 
Independent researcher \and 
Mobile TeleSystems (MTS) \and
Artificial Intelligence Research Institute (AIRI) \\
\email{a.panchenko@skol.tech} 
\\
}

\maketitle   % typeset the header of the contribution
\begin{abstract}
Text style transfer techniques are gaining popularity in natural language processing allowing paraphrasing text in the required form: from toxic to neural, from formal to informal, from old to the modern English language, etc. Solving the task is not sufficient to generate \textit{some} neural/informal/modern text, but it is important to preserve the original content unchanged. This requirement becomes even more critical in some applications such as style transfer of goal-oriented dialogues where the factual information shall be kept to preserve the original message, e.g. ordering a certain type of pizza to a certain address at a certain time. The aspect of content preservation is critical for real-world applications of style transfer studies, but it has received little attention. To bridge this gap we perform a comparison of various style transfer models on the example of the formality transfer domain.  To perform a study of the content preservation abilities of various style transfer methods we create a parallel dataset of formal vs. informal task-oriented dialogues. The key difference between our dataset and the existing ones like GYAFC~\cite{rao-tetreault-2018-dear} is the presence of goal-oriented dialogues with predefined semantic slots essential to be kept during paraphrasing, e.g. named entities. This additional annotation allowed us to conduct a precise comparative study of several state-of-the-art techniques for style transfer. Another result of our study is a modification of the unsupervised method LEWIS \cite{reid-zhong-2021-lewis} which yields a substantial improvement over the original method and all evaluated baselines on the proposed task. 

% or ParaDetox~\cite{logacheva-etal-2022-paradetox}

\end{abstract}

\section{Introduction}
\label{sec:intro}

% Text style transfer (TST) is an NLP task of changing one attribute of text (style) of the source text into the target style keeping the content of the generated text similar to the source text. 

Text style transfer (\textbf{TST}) systems are designed to change the style of the original text to an alternative one, such as more positive~\cite{luo-etal-2019-towards}, more informal~\cite{rao-tetreault-2018-dear}, or even more Shakespearean~\cite{jhamtani-etal-2017-shakespearizing}. Such systems are becoming very popular in the NLP. They could be applied to many purposes: from assistance in writing to diversifying responses of dialogue agents and creating artificial personalities. 

Task-oriented dialogue agents are one of the possible applications of TST. In such dialogues, it is crucial to preserve important information such as product names, addresses, time, etc. Consider the task of making the source sentence from dialogue agent \textit{Do you want to order a pizza to your office at 1760 Polk Street?} more informal to improve the user experience with the agent. This text contains named entities (\textit{pizza}, \textit{1760 Polk Street}) that are critical to understanding the meaning of the query and following the correct scenario and that could be easily lost or corrupted during standard beam search-based generation even if the model is trained on parallel data~\cite{10.1007/978-3-031-08473-7_40}. At the same time, there are several words in this sentence that could be changed to make the style more informal. For example, a target sentence such as \textit{do u wanna order a pizza 2 ur office at 1760 Polk Street?} requires only small edition of some words not related to the important entities. This suggests that it could be better to keep the important entities intact and train the model to fill the gaps between them.
% parallel data.~\cite{10.1007/978-3-031-08473-7_40}

In this work, we focus on text formality transfer, or, more precisely, transferring text style from formal to informal with an additional requirement to preserve the predefined important slots. 
%We assume that the source sentence has a set of predefined slots that should be precisely transferred to the generated text. 
We assume that the transfer task is supervised, which means that a parallel corpus of the text pairs in the source and target style is available (we use the GYAFC dataset~\cite{rao-tetreault-2018-dear}).

% Many types of TST can be performed using small edits of the source text instead of standard text generation from scratch. For example, transferring the formal text into more informal form in many cases could be done by simple editing of verbs (e.g. \textit{want to} to \textit{wanna}) or prepositions (e.g. \textit{to} to \textit{2}). 
A similar intuition has been used in the unsupervised TST domain in LEWIS ~\cite{reid-zhong-2021-lewis}, where the authors created a pseudo-parallel corpus, trained a RoBERTa~\cite{zhuang-etal-2021-robustly} tagger to identify coarse-grain Levenshtein edit types for each token of the original text, and finally used a BART~\cite{lewis-etal-2020-bart} masked language model to infill the final edits. With the increasing interest in the TST field, several large parallel datasets have been collected for the most popular TST directions, such as formality transfer~\cite{rao-tetreault-2018-dear}. Thus, it became possible to use the advantage of parallel data to address the specific task of content preservation.   

The contributions of our work are three-fold:
\begin{enumerate}

\item We present \textit{PreserveDialogue}: the first dataset for evaluating the content-preserving formality transfer model in the task-oriented dialogue domain.
% : the dataset is a parallel formal-informal corpus derived from a dialogue dataset SGDD~\cite{rastogi2020towards}.

\item We perform a study of strong supervised style transfer methods, based on transformer models, such as GPT2 and T5 (as well as simpler baselines), showing that methods based on Levenstein edit distance such as LEWIS~\cite{lewis-etal-2020-bart} are outperforming them if content shall be strictly preserved. 
    
 \item We introduce LEWIT, an improved version of the original LEWIS model based on T5 encoder-decoder trained on parallel data which yields the best results across all tested methods.

\end{enumerate}

We open-source the resulting dataset and the experimental code\footnote{\url{https://github.com/s-nlp/lewit-informal}}. Additionally, we release the best-performing pre-trained model   LEWIT for content preserving formality transfer on HugingFace model hub.\footnote{\url{https://huggingface.co/s-nlp/lewit-informal}} %\footnote{\url{https://huggingface.co/SkolkovoInstitute/LEWIT-informal}} \footnote{Will be published after blind review}%\footnote{\url{https://github.com/skoltech-nlp/LEWIT-informal}} 

% systems are designed to change the style of the original text to alternative one, such as more informal~\cite{rao-tetreault-2018-dear}, more positive~\cite{luo-etal-2019-towards}, or even more Shakespearean~\cite{jhamtani-etal-2017-shakespearizing}.

\section{Related works}

% There are several approaches to content preserving or constrained generation, which have been studied in different fields of NLP.
% не выделять отдельно TST, акцентировать внимание на группах подходов

In this section, we briefly introduce the existing approaches to text generation with an emphasis on preserving certain content.

% The standard approach to text generation is beam search which iteratively generates possible next tokens, and the sequence yielding the highest conditional probability is selected as the best candidate after each iteration. There are several methods to constraint the beam search process which can be roughly divided into two broad categories: hard and soft constraints.

\subsubsection{Constrained beam search}
%  если еокращать объем related works то можно ориентироваться сюда https://aclanthology.org/2020.acl-main.325.pdf

The standard approach to text generation is beam search which iteratively generates possible next tokens, and the sequence yielding the highest conditional probability is selected as the best candidate after each iteration. There are several methods to constraint the beam search process which can be roughly divided into two broad categories: hard and soft constraints. In the hard constrained category, all constraints are ensured to appear in the output sentence, which is generally achieved by the modified type of beam search, allowing to directly specify the tokens to be preserved \cite{hu-etal-2019-improved}. % Works like~\cite{ou-etal-2021-infillmore} yield more flexibility to the form of constraints keeping them as a trie, so each constraint can have several alternative forms. 
Opposite to hard-constrained approaches, soft-constrained approaches modify the model's training process by using the constraints as an auxiliary signal. Such signal is often either marked with special tags~\cite{9108255} or simply replaced with delexicalized tokens~\cite{10.1145/3487351.3490974} during the training process and inference.

\subsubsection{Edit based generation}

Beam search is not the only existing approach to text generation. One popular substitution is Levenstein transformer~\cite{10.5555/3454287.3455290} --- a partially autoregressive encoder-decoder framework based on Transformer architecture~\cite{46201} devised for more flexible and amenable sequence generation. Its decoder models a Markov Decision Process (MDP) that iteratively refines the generated tokens by alternating between the insertion and deletion operations via three classifiers that run sequentially: deletion, a placeholder (predicting the number of tokens to be inserted), and a token classifier.

\subsubsection{Content preservation in text style transfer}

Content preservation in text style transfer has mostly been addressed in the unsupervised domain.
These methods mostly rely on text-editing performed in two steps: using one model to identify the tokens to delete and another model to infill the deleted text slots~\cite{ijcai2019-732}. LEWIS~\cite{reid-zhong-2021-lewis} approach first constructs pseudo parallel corpus using an attention-based detector of style words and two style-specific BART~\cite{lewis-etal-2020-bart} models, then trains a RoBERTa-tagger~\cite{zhuang-etal-2021-robustly} to label the tokens (insert,
replace, delete, keep), and finally fine-tunes style-specific BART masked language models to fill in the slots in the target style. LEWIT %\cite{https://doi.org/10.48550/arxiv.2204.13638} 
extrapolates this idea to a supervised setting. The main features of this work are that the token tagger is trained on tags obtained from parallel data, and the slots are filled with a T5 model~\cite{10.5555/3455716.3455856}, by taking advantage of its initial training task of slot filling.

\section{Datasets}

In this section, we describe the parallel training dataset and the evaluation dataset used respectively for tuning and evaluating the content-preserving formality transfer methods.

\subsection{Parallel training dataset: GYAFC}
\label{sec:train_dataset}

In terms of our work, we assume the availability of parallel data. We also focus our experiments on the transfer of formal text to a more informal form. Grammarly’s Yahoo Answers Formality Corpus (GYAFC) containing over 110K informal/formal sentence pairs fits well to this task. The main topics of the sentences in this dataset are related to either entertainment and music or family and relationships, both of these topics take almost equal part in the dataset~\cite{rao-tetreault-2018-dear}. 

% The initial sentences were scored with a formality classifier trained on PT16 dataset~\cite{pavlick-tetreault-2016-empirical} and then were rewritten in a more formal way and further checked by the annotators hired in Amazon Mechanical Turk.

\begin{figure}[t!]
\centering
 \includegraphics[scale=0.7]{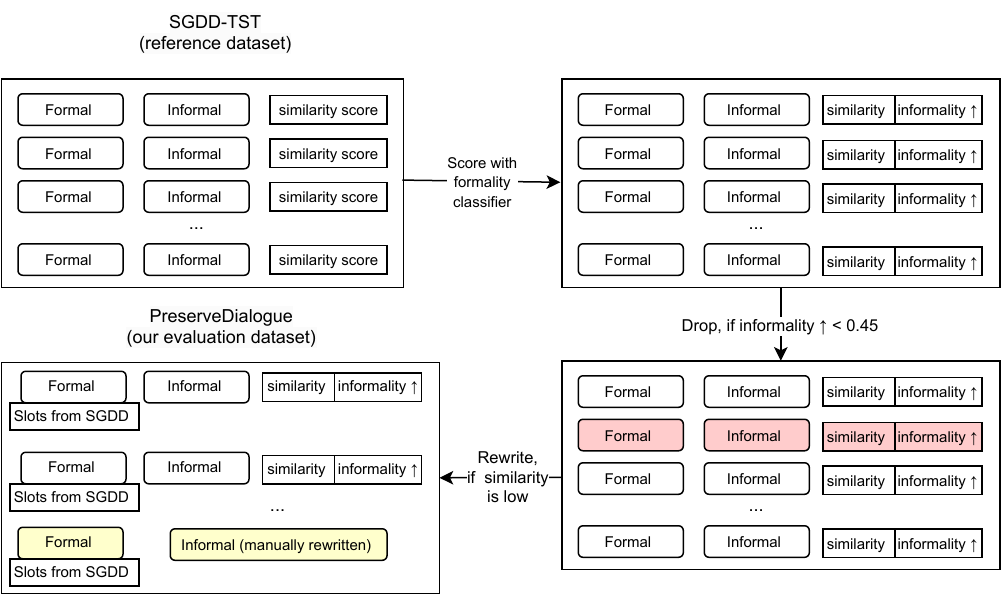}
\caption{The pipeline of collecting the PreserveDialogue dataset. The reference SGDD-TST dataset consists of formal-informal sentence pairs annotated with semantic similarity scores. The pairs are scored with the formality classifier model and the pairs with insignificant informality increase are dropped. The pairs with significant formality increase that have low semantic similarity scores are manually rewritten to be semantically similar. Finally, the important slots  related to the formal sentence are extracted from the SGDD dataset.}
\label{fig:eval_dataset_collection}
\end{figure}

\subsection{Parallel evaluation dataset: PreserveDialogue}
\label{sec:dataset}

% \begin{table}[]
% \small 
% \centering 
% \begin{tabular}{c|c|c}

% \hline 
% & \textbf{\# Samples} & \textbf{Fraction} \\ \hline 
% Total samples &  1100 & 100\% \\ 
% Generated by a model & 731  & 66.5\% \\ 
% Manually rewritten & 369  & 33.5\% \\ \hline 
% \end{tabular}
% \caption{Statistics of the dataset used for evaluation}
% \label{tab:dataset_stat}
% \end{table}

We create a special evaluation dataset denoted as \textit{PreserveDialogue}. It is based on SGDD-TST~\cite{10.1007/978-3-031-08473-7_40}\footnote{\url{https://github.com/s-nlp/SGDD-TST}}. SGDD-TST consists of sentence pairs in a formal and informal style with human annotation of semantic similarity. Its formal phrases were obtained from SGDD~\cite{rastogi2020towards} and informal ones were generated by a large T5-based model tuned on the GYAFC dataset. Some of the generated paraphrases were annotated as semantically different, which is why SGDD-TST in its original form is not appropriate for evaluating content-preserving style transfer. Thus we create PreserveDialogue as a derivative from SGDD-TST. Fig.~\ref{fig:eval_dataset_collection} shows the main steps of the PreserveDialogue collection process. These steps are also described in more details below:
\begin{enumerate}
    \item \textbf{Selecting sentences with significant informality increase.} We score all sentence pairs of SGDD-TST with a formality classifier (described in Section~\ref{sec:style_acc}) and leave only 1100 pairs with an informality increase (namely, the difference between formality classifier scores of formal and informal sentences) greater than the empirically selected threshold of 0.45. 
    \item \textbf{Rewriting paraphrases with the corrupted sense.} Within the 1100 pairs, 369 are not semantically equal to each other (according to the similarity score available in the reference SGDD-TST dataset). Such pairs are  rewritten by the members of our team according to the common intuition of informal style. After the messages are rewritten their informality is verified with the similar formality classifier used in the previous step. 
    \item \textbf{Extracting important slots.} The SGDD dataset~\cite{rastogi2020towards} (the source of formal phrases of SGDD-TST) contains task-oriented dialogues with predefined named entities. We use them as important slots for the sentences in a formal style in PreserveDialogue.
    
    % full of named entities related to real-life tasks (booking of hotels, flights, restaurants). The slots of such entities are explicitly annotated in SGDD. We use the corresponding slots as pre-defined important entities.
\end{enumerate}

% This yields two types of pairs: 731 pairs that have been originally annotated by crowdsource workers as semantically similar (i.e. pre-trained model was able to perform formality transfer without loss of content) and 369 pairs that have been annotated as different or only partially similar (i.e. the model lost the content). The last group of samples was manually rewritten by the members of our team so that the second sentence is semantically similar to the first one. 

Finally, the PreserveDialogue dataset consists of 1100 sentence pairs of formal and informal phrases. By the steps described above we make sure that these pairs have the equivalent sense and significantly differ in terms of informality. Moreover, each pair has a set of entity slots extracted from SGDD~\cite{rastogi2020towards} (predecessor of SGDD-TST). These slots are related to the first sentence in the pair and are considered significant information which should be kept during formality transfer. A sample from the dataset can be found in Table~\ref{tab:dataset_samples}.

\begin{table*}[t!]
\small
\centering
% \begin{tabular}{|l|l|l|}
\begin{tabular}{P{0.35\linewidth}|P{0.2\linewidth}|P{0.35\linewidth}}

\textbf{Source formal text}       & \textbf{Important slots }          & \textbf{Target informal rewrite   }        \\ \hline
Red Joan sounds great   & { Red Joan }          & red joan is cool           \\ \hline
What do I have scheduled Tuesday next week?    & { Tuesday next week } & what stuff do i have to do on tuesday next week? \\ \hline
I am looking for a unisex salon in SFO.       & { SFO }    & i wanna find  a unisex salon in SFO   \\ \hline
Please confirm this: play Are You Ready on TV & { Are You Ready }     & plz confirm this: play Are You Ready on TV       \\ \hline
Where do you want to pick it up at?           & { -- }         & where do u wanna pick it up at?       \\ \hline
\end{tabular}
\caption{Examples of texts from the PreserveDialogue dataset used for evaluation.}
\label{tab:dataset_samples}
\end{table*}

% https://disk.yandex.ru/client/disk/work%20backup/TST_backup/tst_eval - some code saved on Yandex disk
\section{Evaluation}
\label{sec:evaluation}
In this section, we describe the methods of automatic evaluation of the formality transfer models.

In most TST papers (e.g.~\cite{zhang-etal-2020-parallel,moskovskiy-etal-2022-exploring,NIPS2017_2d2c8394}) the methods are evaluated with the combination of the measures that score the basic TST properties: style accuracy, content preservation, and fluency. Our work is dedicated to formality transfer with an additional task to preserve particular slots. That is why we have to use an additional evaluation method: slot preservation. All of these measures evaluate TST quality from significantly different points of view, thus to reveal the TST method that performs best we aggregate them by \textit{multiplying the four measures for each sentence and then averaging the products over the dataset}, following the logic of \cite{krishna-etal-2020-reformulating}. More details about each measure are provided in the following paragraphs. The code of measures calculations is also open-sourced in our repository.

% NIPS2017_2d2c8394, NEURIPS2019_8804f94e,mti5090054

% \begin{itemize}
%     \item Content preservation. The score of semantic similarity between source and genertaed text,
%     \item Fluency. The score of naturalness of the generated text.
%     \item Style accuracy. The score showing how well the generated text corresponds to the target style.
% \end{itemize}

% As stated in Section~\ref{sec:general_tst_def} our work is dedicated to transferring from formal to informal style. 
% Our preliminary studies showed that neither of the named fluency calculation approaches performs well with informal texts because the nature of informality itself means that these texts are in some sense uncommon. Thus we intentionally skip this measure.
% Instead of language quality scoring, we pay additional attention to content preservation by calculating the percentage of slots (described in Sections~\ref{sec:general_tst_def} and  \ref{sec:dataset}) from the original texts preserved in the generated texts. 

\subsubsection{Content preservation} 

% \subsection{Content preservation}
% \label{sec:content}

To score the general similarity between the generated text and the reference informal text we use Mutual Implication Score\footnote{\url{https://huggingface.co/s-nlp/Mutual-Implication-Score}}, a measure of content preservation based on predictions of NLI models in two directions. This measure has been compared~\cite{babakov-etal-2022-large} to a large number of SOTA content similarity measures and it was shown that it demonstrates one of the highest correlations with human judgments in the formality transfer domain: Spearman correlation between 0.62 and 0.77 depending on the dataset. 
% The score is calculated between target informal rewrites and the rewrites generated with a model.
% ~\cite{babakov-etal-2022-large}

\subsubsection{Slots preservation}

The key point of content preservation, especially in task-oriented dialogues, is keeping the important entities from a source sentence (see Section~\ref{sec:dataset}). Thus, we check whether these entities exist in the generated sentence. Most entities could have at least two different forms, which could be considered correct (e.g. ``fourteen'' and ``14''). To ensure that the entity is not considered lost even if it is generated in an alternative form, we normalize the important slots and generated text in the following way. All text tokens are lowercased and lemmatized. The state names (e.g. ``Los Angeles''---``LA''), numbers (e.g. ``six''---``6''), and time values (e.g. ``9am''---``9a.m.''---``nine in the morning'') are adjusted to a standard form using a set of rules. Some frequent abbreviations (geographic entity types, currencies) are expanded. The slots that still were not matched exactly are matched to the n-gram of the new sentence with the highest ChrF score~\cite{popovic-2015-chrf}. The final slots preservation score is calculated as the ratio of the preserved slots in a new sentence (with ChrF scores as weights for the slots that were matched approximately) to the total number of slots in a source sentence. This ratio calculation takes into account both original and standardized forms of the tokens. This approach uses the idea similar to copy success rate calculation used for scoring constraints preservation in machine translation~\cite{ijcai2020-0496}.

\subsubsection{Style accuracy}
\label{sec:style_acc}
% описать модель стиля так как на нее была ссылка ранее

To ensure that the generated text corresponds to the target style we use a RoBERTA-based formality ranker\footnote{\url{https://huggingface.co/s-nlp/roberta-base-formality-ranker}}.
 The ranker was trained on two formality datasets: GYAFC~\cite{rao-tetreault-2018-dear} and P\&T~\cite{pavlick-tetreault-2016-empirical}. We verified the quality of this ranker by calculating the Spearman correlation of its score on the test split of GYAFC and P\&T, which was 0.82 and 0.76 correspondingly.

\subsubsection{Fluency}
\label{sec:style_acc}

The generated text should look natural, grammatical and fluent. Fluency is often evaluated as the perplexity of a large language model, but to make the results more interpretable, we use a RoBERTA-based classifier trained on the Corpus of Linguistic Acceptability (CoLA)~\cite{warstadt2018neural}. It is a diverse dataset of English sentences annotated in a binary way, as grammatically acceptable or not. A detailed justification of using a CoLA-based classifier for fluency evaluation is presented in \cite{krishna-etal-2020-reformulating}. We use an opensource RoBERTA-based classifier\footnote{https://huggingface.co/textattack/roberta-base-CoLA} trained on CoLA for 5 epochs with a batch size of 32, a learning rate of 2e-05, and a maximum sequence length of 128. Its scores range from 0 to 1 with greater values meaning higher quality, just like all the other metrics we use for evaluation. The reported accuracy of this model on the CoLA validation set is 0.85.
% CoLA

% We choose this method over perplexity, because it ranges from 0 to 1 and its greater values mean
% higher quality, just like metrics we use for evaluating toxicity and content

% Fluency or language quality is typically calculated with a linguistic-acceptability classifier based on such datasets as COLA~\cite{warstadt2018neural} or with the perplexity of a large language model. 

% \subsection{Discussion}

% It is important to consider the fact that automatic measures, such as those presented above, have their limitations. The content preservation measure used in our study (MIS) was proven to correlate well with human judgments in previous works~\cite{babakov-etal-2022-large}, formality ranker was trained on two relevant datasets and showed reasonable metrics on test splits of these datasets. However, in both of these cases, these measures still have room for improvement. At the same time, the slots preservation measure is intuitive but relies a lot on manually designed rules for matching different forms of similar entities, which also can cover only a limited number of cases. The most reliable way to assess the quality of the introduced approach is a manual evaluation yet it requires substantial human labor and is not reproducible. Thus, we settle with the current setup.

% Finally, the skipping of fluency measure is a step that is worth further studying in terms of formality transfer. T

\section{Supervised Style Transfer Methods}

In this section, we describe the baseline methods used in our computational study dedicated to finding the best approach to content-preserving formality transfer. 
%
%The task stated in Section~\ref{sec:general_tst_def} assumes that the important slots are available for the inference of the model. However, this is certainly not always the case, which is why some of the methods assume the availability of the predefined important slots and others do not. These two groups of methods are defined together in the following parapgraphs but are further compared to each other separately in Section~\ref{sec:results}.
%
All models requiring tuning described in this section are tuned on the GYAFC parallel dataset (see Section~\ref{sec:train_dataset}).

\subsection{Naive baselines}

We use two naive baselines. In \textit{copy-paste}, we simply copying the source text, and in \textit{only-slots}, the target string is simply a list of important slots separated by commas. The motivation of these methods is the sanity check of the proposed evaluation pipeline (see Section~\ref{sec:evaluation}). The joint score (multiplication of four measures) is supposed to place the naive methods at the bottom of the leaderboard, which could be treated as a necessary condition of acceptance of the proposed evaluation method.

\subsection{Sequence-to-sequence approaches}

As the setting of our work assumes the availability of parallel data, it is natural to try standard sequence-to-sequence models (seq2seq), both ``as is'' and with some  modifications related to the task of content and slots preservation.

\subsubsection{Standard seq2seq}

We tune the following models in the standard seq2seq approach: pure T5-base (\textit{seq2seq-t5}) and T5-base pre-tuned on a paraphrasing datasets\footnote{\url{https://huggingface.co/ceshine/t5-paraphrase-paws-msrp-opinosis}} (\textit{seq2seq-t5-para}). We also experiment with using a template generated from the target sentence as a text pair for the training of the T5 model (\textit{seq2seq-t5-template}).

% .and with using lexically constrained beam search  (\textit{seq2seq-t5-para-constr}, \textit{seq2seq-t5-constr}). It is mainly based on dynamic beam allocation and is implemented in HuggingFace library\footnote{\url{https://github.com/huggingface/transformers/issues/14081}}. 

\subsubsection{seq2seq with hard lexical constraints}

The models trained in the standard seq2seq approach can be inferenced with lexically constrained beam search 
(\textit{seq2seq-t5-para-constr}, \textit{seq2seq-t5-constr}) which is implemented in the HuggingFace library mainly based on dynamic beam allocation\footnote{\url{https://github.com/huggingface/transformers/issues/14081}}~\cite{hu-etal-2019-improved}.

\subsubsection{Re-ranking beam search outputs with a neural textual similarity metric.}

We experiment with re-ranking beam search outputs with neural textual similarity metric. The hypothesis obtained after the beam search could be re-ranked w.r.t. some content preservation measure. 
To avoid overfitting, we should not rerank with the same measure (MIS) that we use for evaluation.   In \cite{babakov-etal-2022-large}, the authors show that apart from MIS, BLEURT~\cite{sellam-etal-2020-bleurt} also demonstrates reasonable performance in the formality transfer domain. We use a mean of BLEURT-score and conditional probability to perform a final re-ranking of the hypothesis generated after the beam search. This approach is used in combination with seq2seq-para-constr (\textit{rerank-BLEURT-constr}) and with seq2seq-para\ (\textit{rerank-BLEURT}).
% \cite{babakov-etal-2022-large} showed that apart of MIS used as one of the evaluation methods for our study,

\subsubsection{Learning to preserve slots with tags}

Finally, we try to embed the task of content preservation into the seq2seq training. One of the possible ways to do that is to embed a signal in the training data indicating that a certain slot should be preserved. We use two different types of such signals. First, similarly to the idea presented in~\cite{zhang-etal-2021-dont} we put special \texttt{<tag>} tokens around the slots to be preserved (\textit{slot-tags}). Second, we replace the whole slot with a placeholder token and train model to re-generate this placeholder, which is then filled with the value from the original sentence~\cite{10.1145/3487351.3490974} (\textit{delex}).

\subsection{Language models inference}

There exists some evidence of the possibility to use the large pre-trained language models (LM) in zero- and few-shot way~\cite{brown2020language}. The LM can also be slightly fine-tuned on the parallel data to be capable of performing the desired task.

% It is necessary to either choose the proper prompt making a language model to perform a desired task, or to prepend some training samples to this prompt. 
% These approaches are named zero and few shot learning. 

Similarly to the idea of~\cite{reif-etal-2022-recipe}, we construct a prompt for the language model to make it generate more informal text: \textit{``Here is a text, which is formal: $<$formal text$>$. Here is a rewrite of the text which contains $<$slot 1$>$, $<$slot 2$>$ and is more informal''} and train  GPT2-medium\footnote{\url{https://huggingface.co/gpt2-medium}} on parallel data to continue this prompt. We use two variations of such approach: with (\textit{GPT2-tuned-constr}) and without (\textit{GPT2-tuned}) the information about constraints in the prompt.

\subsection{LEWIS and its modifications}
% основное отличие что движемся не потокенно а от задачи MLM нативной для Т5

% In this section we introduce the main method ??? Understand what to write after discussion

An intuitively straightforward approach for a human to generate an informal paraphrase is to apply some slight modifications to the formal source text. This group of approaches is named ``edit-based''. Most of these approaches use numerous models to perform separate edition actions for generating a new text: deletion, insertion, placeholder, infiller models~\cite{zhang-etal-2020-pointer}. 
We experimented with the LEWIS model~\cite{reid-zhong-2021-lewis} representing this kind of methods. 

LEWIS was designed in the unsupervised domain, so the authors first created a pseudo-parallel corpus, then trained a RoBERTa tagger to identify coarse-grain Levenshtein edit types for each token from the original text, and finally used a BART masked language model to infill the final edits. We use LEWIS in a parallel data setting by tuning BART on our parallel dataset and using it either with known constraints (\textit{LEWIS-constr}) or with the labels inferred from the RoBERTa tagger trained on the edits from parallel data (\textit{LEWIS-tag}).

\begin{figure*}[t!]
\centering
%  \makebox[\textwidth][c]{\includegraphics[scale=0.47]{media/paraphrase_metrics.pdf}}%
%  \hspace*{-1.cm}{\includegraphics[scale=0.5]{media/inference.pdf}}
 \includegraphics[scale=0.5]{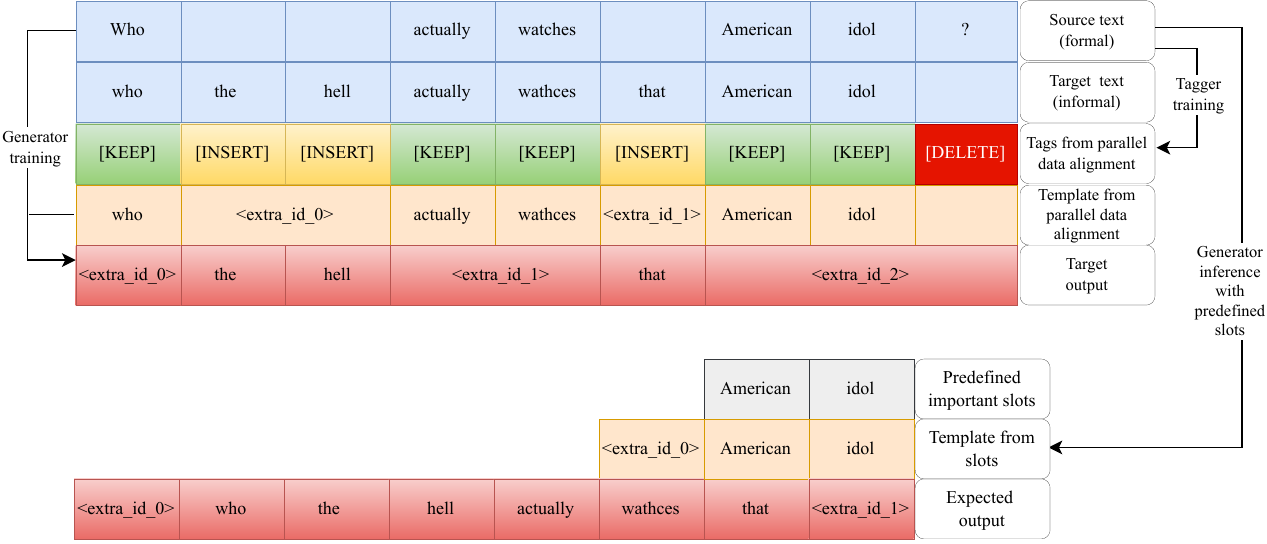}
 \caption{LEWIT model workflow: The edit tags are obtained from the alignment of source formal and target informal texts. These tags are used to train the token tagger. These edit tags are also used to create a template used to train a T5-generator model, which fills the slots between the preserved tokens. %If the predefined important slots are known they can be used to create a template used for inference of the pre-trained generator model.
 }
\label{fig:LEWIT_scheme}
\end{figure*}

% similar to the ones used for LEWIT-tag (\textit{LEWIS-tag}).

We also test a modified version of LEWIS denoted as LEWIT (T5-based LEWIS): similarly to LEWIS architecture it involves a token tagger trained on Levenshtein edits obtained from the alignment of the parallel data, but its infiller model (i.e. the model inserting the tokens between other tokens) is based on T5 that was originally trained with the specific task of infilling gaps of several tokens.
%
% susanto-etal-2020-lexically
%
%LEWIT approach was originally introduced with text detoxification task and proved to demonstrate a significant performance on it\footnote{Will be published after blind review}. The initial setting of this approach seems very suitable for content preserving formality transfer task, that is why we introduce it in our work and use it as the main hypothesis.
%
%LEWIT aims to combine the advantages of edit-based operations and parallel data availability. 
The LEWIT model consists of two steps as illustrated in Figure \ref{fig:LEWIT_scheme}. First, the RoBERTa token tagger is trained on the tags from the GYAFC dataset (see Section~\ref{sec:train_dataset}) which are directly computed from edits required to transform the source texts into the target texts. Second, the T5-based\footnote{\url{https://huggingface.co/t5-base}} generator model is trained on the templates from parallel data that was also generated from the GYAFC dataset sentence pairs. The model receives the concatenation of the source sentence and the template constructed w.r.t. the edit tokens and is expected to generate the words masked from the target sentence. 

LEWIT inherits the general logic of LEWIS. Its distinguishing feature is that its generator model is T5-based. The choice of this model seems more suitable for this task, because gap filling is the main pre-training objective of T5, whereas BART has been pretrained to reconstruct texts with many other types of noise, such as token deletion, sentence permutation, and document rotation.
% that skips the necessity of using multiple models for an edit-based generation. 

In terms of the content preserving formality transfer task, the important slots are sent to the model together with the source text. As we assume the availability of the parallel data, we get the list of important slots from the words of the target text that are similar to the ones in the source text.  However, the first part of the LEWIT pipeline (token tagger) can also generate labels indicating which tokens should be preserved. Thus, we try combinations of the templates used for inference of the trained generator model: from predefined slots only, like shown on the bottom part of Figure~\ref{fig:LEWIT_scheme}, (\textit{LEWIT-constr}) and from predefined slots and tagger labels (\textit{LEWIT-constr-tag}).

We perform additional experiments that do not assume the availability of predefined slots. The templates for these approaches are obtained from the aforementioned RoBERTa-based tagger labels (\textit{LEWIT-tag}) and third-party NER-tagger (\textit{LEWIT-NER}). A significant part of the tags generated with the tagger within the test set was either ``replace'' (46,7\%) or ``equal'' (50\%). ``Delete'' and ``insert'' took 3\% and 0.3\%  correspondingly. This  proportion corresponds to the general intuition of small-edits-based paraphrasing of formal texts into more informal style by keeping the most important content intact and either slightly altering or sometimes deleting less important parts.

% e\footnote{\url{https://huggingface.co/facebook/bart-base}}

% The most direct way is to fine-tune a sequence-to-sequence model on a parallel corpus. Inputs are
% templates from the tagger, and outputs are masked words from the target sentence. The whole process
% of data generation for training is in Figure 1. We also concatenate a source sentence with the generated
% template, as in Reid and Zhong (2021), to provide access to the original masked words.

\section{Results}
\label{sec:results}

\begin{table*}[h!]
% \small
\centering
% \begin{tabular}{ccccc} 
\resizebox{\textwidth}{!}{
\begin{tabular}{l|c|c|c|c|c}

\textbf{Method} & \textbf{Style Accuracy} & \textbf{Content Preservation} & \textbf{Slot Preservation} & \textbf{Fluency} & \textbf{Joint} \\ \hline

\multicolumn{6}{l}{\textit{Without known constraints}} \\ \hline

LEWIT-tag (T5)  & 0.69 & 0.85 & 0.98 & 0.75* & \textbf{0.43} \\
LEWIT-NER (T5)           & 0.82 & 0.74 & 0.94 & 0.74 & 0.42 \\
LEWIS-tag~\cite{reid-zhong-2021-lewis} (BART)  & 0.77 & 0.69 & 0.99 & 0.72 & 0.38 \\
seq2seq-t5-para     & 0.60  & 0.82 & 0.95 & 0.74 & 0.34 \\
seq2seq-t5           & 0.54 & 0.87 & 0.98 & 0.73 & 0.33 \\
rerank-BLEURT        & 0.46 & 0.84 & 0.97 & 0.75* & 0.28 \\
GPT2-tuned~\cite{reif-etal-2022-recipe} & \textbf{0.91} & 0.57 & 0.76 & 0.69 & 0.27 \\
copy-paste & 0.03 & \textbf{0.90}  & \textbf{1.00}    & \textbf{0.83} & 0.02\\ \hline

\multicolumn{6}{l}{\textit{With known constraints}} \\ \hline
LEWIT-constr (T5)       & 0.80  & 0.76 & 1.00    & 0.76* & \textbf{0.46} \\
LEWIT-constr-tag  (T5)     & 0.73 & \textbf{0.83} & 1.00    & 0.74 & 0.45 \\
LEWIS-constr~\cite{reid-zhong-2021-lewis} (BART)  & 0.81 & 0.68 & 1.00 & 0.75 & 0.41 \\
delex~\cite{10.1145/3487351.3490974} & 0.69 & 0.75 & 0.98 & 0.78* & 0.40  \\
seq2seq-t5-template & 0.49 & 0.83 & 0.97 & 0.78* & 0.31 \\
seq2seq-t5-constr~\cite{hu-etal-2019-improved}       & 0.71 & 0.71 & 1.00    & 0.61 & 0.31 \\
slot-tags~\cite{zhang-etal-2021-dont}    & 0.56 & 0.74 & 0.97 & 0.76 & 0.31 \\
seq2seq-t5-para-constr & 0.64 & 0.74 & 1.00    & 0.61 & 0.29 \\
rerank-BLEURT-constr    & 0.54 & 0.77 & 1.00    & 0.64 & 0.27 \\
GPT2-tuned-constr~\cite{reif-etal-2022-recipe}        & \textbf{0.91} & 0.41 & 0.82 & 0.68 & 0.21 \\
only-slots    & 0.73 & 0.21 & 1.00    & \textbf{0.81} & 0.12 \\ \hline
\end{tabular}
}
\caption{Results with and without the usage of the predefined important slots. ``Joint'' is the average product of all four measures. The
values \textbf{in bold} show the highest value of the metric with the significance level of $\alpha$ = 0.05 (by Wilcoxon signed-rank test). The values with an insignificant difference between LEWIT and other methods are marked with a ``*'' sign. The highest value for the slot preservation for the methods with known constraints is not indicated because in most cases all constraints are preserved by the design of the methods from this group.}
\label{tab:leaderboard}
\end{table*}

% \begin{tabular}{P{0.5\linewidth}P{0.1\linewidth}P{0.1\linewidth}P{0.1\linewidth}P{0.1\linewidth}}

The results are grouped according to the availability of the predefined important slots or constraints in the inference time and are shown in Table~\ref{tab:leaderboard}. We can see that both naive approaches are pushed to the bottom of the tables and their joint measure value is substantially less than the closest non-naive approach. %This provides an additional guarantee that the chosen evaluation method is reliable. %
We can also see that the LEWIT approach outperforms all strong baselines in both settings of the experiments. 

% The examples of formality paraphrases generated by the top-performing systems are shown in Appendix in Tables~\ref{tab:appendix_constr_samples} and \ref{tab:appendix_no_constr_samples}.

\textbf{Case 1: the slots are not known in the inference time.} Both NER-tagger and edits-tagger-based approaches perform similarly by the joint measure outperforming the baseline methods. The edits-tagger approach yields better content and slots preservation but worse style transfer accuracy. The examples of the generated paraphrases are shown in Table~\ref{tab:appendix_constr_samples}.

\textbf{Case 2: the slots are known in the inference time.} Different variations of LEWIT also outperform the baseline methods. We can see that if the important slots are known, their combination with the edits token tagger can increase content preservation, however, this yields a decrease in style accuracy. The examples of the generated paraphrases are shown in Table~\ref{tab:appendix_no_constr_samples}.

\begin{table*}[]
\small
%\centering
% \begin{tabular}{|l|l|l|l|}
\begin{tabular}{P{0.30\linewidth}|P{0.15\linewidth}|P{0.40\linewidth}|P{0.15\linewidth}}
\hline
\textbf{Formal source text }  & \textbf{Important slots }          & \textbf{Informal rewrite }          & \textbf{System}        \\ \hline
\multirow{3}{=}{\centering SAN International Airport is the location of flight departure.}   
& \multirow{3}{=}{SAN International Airport}           & SAN International Airport is where the flight departs       & LEWIT-constr \\ \cline{3-4} 
   &     & SAN International Airport International airport is the place of flight departure.          & LEWIS-constr \\ \cline{3-4} 
    &     & SAN International Airport is the start of your flight.      & delex         \\ \hline
% \multirowcell{Okay. You'll instead book a table for 1 at 54 Mint Ristorante Italiano in San Francisco at 1:15 pm on March 14th?} & \multirowcell{54 Mint Ristorante Italiano; San Francisco; 1:15 pm; March 14th} & ok u'll instead book 1 table at 54 Mint Ristorante Italiano in San Francisco ( 1:15 pm, March 14th )    & LEWIT-constr \\ \cline{3-4} 
%           &     & u'll instead book a table for 1 at 54 mint ristorante Italiano in san francisco 1:15 pm March 14th      & LEWIS-constr \\ \cline{3-4} 
%           &     & Okay, you'll just go to 54 Mint Ristorante Italiano in San Francisco and get a seat for 1 at March 14th on 1:15 pm & delex         \\ \hline
\multirow{3}{=}{\centering How will the weather be in Delhi, India on the tomorrow?}   & \multirow{3}{=}{Delhi; India; tomorrow}         & whats the weather like in Delhi, India for tomorrow .       & LEWIT-constr \\ \cline{3-4} 
     &        & Will the weather be in india on the next day?    & LEWIS-constr \\ \cline{3-4} 
          &     & i'm not sure, but it's going to be cool in Delhi, India on tomorrow    & delex         \\ \hline
\multirow{3}{=}{\centering Your destination will be at Sacramento Valley Station.}   & \multirow{3}{=}{Sacramento Valley Station}         & go to Sacramento Valley Station !       & LEWIT-constr \\ \cline{3-4} 
          &     & destination will be at Sacramento Valley Station    & LEWIS-constr \\ \cline{3-4} 
          &     & Sacramento Valley Station    & delex         \\ 
\hline

\end{tabular}
\caption{Examples of samples generated by top performing formality transfer systems with known constraints.}
\label{tab:appendix_constr_samples}
\end{table*}

\begin{table*}[t!]
\small
%\centering
\begin{tabular}{P{0.35\linewidth}|P{0.45\linewidth}|P{0.2\linewidth}}
\hline
\textbf{Formal source text }    & \textbf{Informal rewrite }         & \textbf{System}        \\ \hline
\multirow{3}{=}{\centering No I am leaving on the 3rd from Seattle, WA.}      & No I 'm leaving on the 3rd from Seattle, WA     & LEWIT-tag    \\ \cline{2-3} 
 & No I leaving on the 3rd from Seattle, WA        & LEWIS-tag    \\ \cline{2-3} 
 & No I'm leaving on the 3rd from Seattle WA.      & seq2seq-t5-para \\ \hline
\multirow{3}{=}{\centering Do you have any preference in city and type of events, for example, music or Sports something like that?} & do you like city or music or Sports something like that?   & LEWIT-tag    \\ \cline{2-3} 
 & u like city and sports like that music or sports something like that? & LEWIS-tag    \\ \cline{2-3} 
 & Do you like city and type of things, like music or sports? & seq2seq-t5-para \\ \hline
\multirow{3}{=}{\centering I will be returning Tuesday next week.} & I 'll be back Tuesday next week !    & LEWIT-tag    \\ \cline{2-3} 
 & I l be back tUESDAY next week.       & LEWIS-tag    \\ \cline{2-3} 
 & I'll be back on Tuesday next week.   & seq2seq-t5-para \\ \hline
\multirow{3}{=}{\centering I would like to leave tomorrow from Atlanta.} & I 'm leaving tomorrow from Atlanta.    & LEWIT-tag    \\ \cline{2-3} 
 & I to get away from Atlanta.tomorrow & LEWIS-tag    \\ \cline{2-3} 
 & i want to leave tomorrow from atlanta   & seq2seq-t5-para \\ \hline
% \multirow{3}{=}{\centering Your destination will be at Sacramento Valley Station.} & you'll end up at Sacramento Valley Station !   & LEWIT-tag    \\ \cline{2-3} 
%  & destination will be at Sacramento Valley Station       & LEWIS-tag    \\ \cline{2-3} 
%  & your destination will be at california valley station.  & seq2seq-t5-para \\ \hline
% \multirowcell{I will depart from New Yourk at around quarter past 2 in the afternoon on 9th of March.} & I 'm going to New Yourk at 2 in the afternoon ( 9th of March )   & LEWIT-tag    \\ \cline{2-3} 
%  & i'll depart from ny at 2 on 9th of March.     & LEWIS-tag    \\ \cline{2-3} 
%  & i'll go out from new yourk about quarter past 2 in the afternoon on 9th of march.   & seq2seq-t5-para \\ \hline
% \multirowcell{I want a General Practitioner. Try searching in Lafayette.} & I want a General Practitioner, Try looking in Lafayette !    & LEWIT-tag    \\ \cline{2-3} 
%  & i want a general practitioner. try in lafayette           & LEWIS-tag    \\ \cline{2-3} 
%  & i want a general practitioner try Lafayette.  & seq2seq-t5-para \\ \hline
\end{tabular}
\caption{Examples of samples generated by top performing formality transfer systems without known constraints.}
\label{tab:appendix_no_constr_samples}
\end{table*}

The examples suggest that pure seq2seq models (such as \textit{seq2seq-t5-para}) occasionally change the overall intent or specific slots in undesirable ways, and simple approaches to slot preservation (such as \textit{delex}) sometimes result in unnatural outputs. Edit-based methods seem to avoid these problems in most cases.

LEWIS-based models demonstrate top performance within the baselines, however, LEWIT still performs better according to the joint score. This is most probably an evidence that T5 fits better than the BART model for the specific task of gap filling.

Results in Table~\ref{tab:leaderboard} pass Wilcoxon signed-rank test~\cite{10.2307/3001968} with a significance threshold of 0.05. We tested the hypothesis of the significance of the difference between the best-performing LEWIT method within each group and all baseline methods. The test was performed by splitting the test set into 30 random parts of 900 samples and calculating the significance over the mean of the measurement values from the selected samples. 

% LEWIT was previously tested on detoxification task in terms of text detoxification based on parallel corpora workshop\footnote{Will be published after review}, where the participants were provided with a parallel dataset of toxic and neutral utterances. The solution based on LEWIT took fourth place according to the human evaluation. More details about this competition could be found in Appendix~\ref{sec:LEWIT_detox}. 
% https://russe.nlpub.org/2022/tox/

Certainly, LEWIT has its limitations. The most notable one is that sticking to the structure of the source sentence limits the ability of the TST model to alter its syntactical structure. Moreover, in some contexts, a text may look more natural if rewritten from scratch. 

We see the main use-case of LEWIT in applications where exact preservation of content is crucial, such as goal-oriented dialogue systems (e.g. pizza ordering), where communication goals cannot be compromised for better fluency. 

\section{Conclusions}

In this paper, we study the ways of supervised transfer of formal text to more informal paraphrases with special attention to preserving the content. In this task, the content of the source text is supposed to have a set of predefined important slots that should be kept in the generated text in either their original or slightly changed form but without a change of their meaning. To evaluate various methods for this task we collect a dataset of parallel formal-informal texts all of which have a set of predefined important slots. Using the new dataset we perform a computational study of modern approaches to supervised style transfer in two settings: with and without information about the predefined important slots provided at the inference time. 

Results of our study show that if content preservation is a crucial goal, methods that do not rewrite the text completely are preferable. In this setting, it is better to use a token tagger marking spans with key information to be kept (named entities, etc.) from everything else which can be rewritten more freely with a separate generator that rephrases the rest. We show that the LEWIS~\cite{lewis-etal-2020-bart} approach operating in this way outperforms strong baselines trained on parallel data by a large margin. We also show the original model can be substantially further improved if the T5-based generator is used. 

% The examples of formality paraphrases generated by the top performing systems are shown in Appendix in Tables~\ref{tab:appendix_constr_samples} and \ref{tab:appendix_no_constr_samples}.

% \section{Limitations} 
% (i) To apply our approach a set of lexical constrains (e.g. semantic slots) should be provided or some tagger should be used. This cause an additional computational overhead.

% (ii) Not all texts contain entities or some semantic slots to be preserved. Not all application require to optimize for content preservation and the conventional seq2seq may be just enough.

\section*{Acknowledgements}

This work was supported by the MTS-Skoltech laboratory on AI, the European Union's Horizon 2020 research and innovation program under the Marie Skodowska-Curie grant agreement No 860621, and the Galician Ministry of Culture, Education, Professional Training, and University and the European Regional Development Fund (ERDF/FEDER program) under grant ED431G2019/04.

\bibliographystyle{splncs04}
\bibliography{custom}

\begin{thebibliography}{10}
\providecommand{\url}[1]{\texttt{#1}}
\providecommand{\urlprefix}{URL }
\providecommand{\doi}[1]{https://doi.org/#1}

\bibitem{10.1007/978-3-031-08473-7_40}
Babakov, N., Dale, D., Logacheva, V., Krotova, I., Panchenko, A.: Studying
  the role of named entities for content preservation in text style
  transfer. In: Rosso, P., Basile, V., Mart{\'i}nez, R., M{\'e}tais, E.,
  Meziane, F. (eds.) Natural Language Processing and Information Systems. pp.
  437--448. Springer International Publishing, Cham (2022)

\bibitem{babakov-etal-2022-large}
Babakov, N., Dale, D., Logacheva, V., Panchenko, A.: A large-scale
  computational study of content preservation measures for text style transfer
  and paraphrase generation. In: Proceedings of the 60th Annual Meeting of the
  Association for Computational Linguistics: Student Research Workshop. pp.
  300--321. Association for Computational Linguistics, Dublin, Ireland (May
  2022)

\bibitem{brown2020language}
Brown, T., Mann, B., Ryder, N., Subbiah, M., Kaplan, J.D., Dhariwal, P.,
  Neelakantan, A., Shyam, P., Sastry, G., Askell, A., et~al.: Language models
  are few-shot learners. Advances in neural information processing systems
  \textbf{33},  1877--1901 (2020)

\bibitem{ijcai2020-0496}
Chen, G., Chen, Y., Wang, Y., Li, V.O.: Lexical-constraint-aware neural machine
  translation via data augmentation. In: Bessiere, C. (ed.) Proceedings of the
  Twenty-Ninth International Joint Conference on Artificial Intelligence,
  {IJCAI-20}. pp. 3587--3593. International Joint Conferences on Artificial
  Intelligence Organization (7 2020), main track

\bibitem{10.1145/3487351.3490974}
Cui, R., Agrawal, G., Ramnath, R.: Constraint-embedded paraphrase generation
  for commercial tweets. In: Proceedings of the 2021 IEEE/ACM International
  Conference on Advances in Social Networks Analysis and Mining. p. 369–376.
  ASONAM '21, Association for Computing Machinery, New York, NY, USA (2021)

\bibitem{10.5555/3454287.3455290}
Gu, J., Wang, C., Junbo, J.Z.: Levenshtein Transformer. Curran Associates Inc.,
  Red Hook, NY, USA (2019)

\bibitem{hu-etal-2019-improved}
Hu, J.E., Khayrallah, H., Culkin, R., Xia, P., Chen, T., Post, M., Van~Durme,
  B.: Improved lexically constrained decoding for translation and monolingual
  rewriting. In: Proceedings of the 2019 Conference of the North {A}merican
  Chapter of the Association for Computational Linguistics: Human Language
  Technologies, Volume 1 (Long and Short Papers). pp. 839--850. Association for
  Computational Linguistics, Minneapolis, Minnesota (Jun 2019)

\bibitem{jhamtani-etal-2017-shakespearizing}
Jhamtani, H., Gangal, V., Hovy, E., Nyberg, E.: Shakespearizing modern language
  using copy-enriched sequence to sequence models. In: Proceedings of the
  Workshop on Stylistic Variation. pp. 10--19. Association for Computational
  Linguistics, Copenhagen, Denmark (2017)

\bibitem{krishna-etal-2020-reformulating}
Krishna, K., Wieting, J., Iyyer, M.: Reformulating unsupervised style transfer
  as paraphrase generation. In: Proceedings of the 2020 Conference on Empirical
  Methods in Natural Language Processing (EMNLP). pp. 737--762. Association for
  Computational Linguistics, Online (Nov 2020).
  \doi{10.18653/v1/2020.emnlp-main.55}

\bibitem{lewis-etal-2020-bart}
Lewis, M., Liu, Y., Goyal, N., Ghazvininejad, M., Mohamed, A., Levy, O.,
  Stoyanov, V., Zettlemoyer, L.: {BART}: Denoising sequence-to-sequence
  pre-training for natural language generation, translation, and comprehension.
  In: Proceedings of the 58th Annual Meeting of the Association for
  Computational Linguistics. pp. 7871--7880. Association for Computational
  Linguistics, Online (Jul 2020)

\bibitem{9108255}
Li, H., Huang, G., Cai, D., Liu, L.: Neural machine translation with noisy
  lexical constraints. IEEE/ACM Transactions on Audio, Speech, and Language
  Processing  \textbf{28},  1864--1874 (2020). \doi{10.1109/TASLP.2020.2999724}

\bibitem{luo-etal-2019-towards}
Luo, F., Li, P., Yang, P., Zhou, J., Tan, Y., Chang, B., Sui, Z., Sun, X.:
  Towards fine-grained text sentiment transfer. In: Proceedings of the 57th
  Annual Meeting of the Association for Computational Linguistics. pp.
  2013--2022. Association for Computational Linguistics, Florence, Italy (Jul
  2019)

\bibitem{moskovskiy-etal-2022-exploring}
Moskovskiy, D., Dementieva, D., Panchenko, A.: Exploring cross-lingual text
  detoxification with large multilingual language models. In: Proceedings of
  the 60th Annual Meeting of the Association for Computational Linguistics:
  Student Research Workshop. pp. 346--354. Association for Computational
  Linguistics, Dublin, Ireland (May 2022)

\bibitem{pavlick-tetreault-2016-empirical}
Pavlick, E., Tetreault, J.: An empirical analysis of formality in online
  communication. Transactions of the Association for Computational Linguistics
  \textbf{4},  61--74 (2016)

\bibitem{popovic-2015-chrf}
Popovi{\'c}, M.: chr{F}: character n-gram {F}-score for automatic {MT}
  evaluation. In: Proceedings of the Tenth Workshop on Statistical Machine
  Translation. pp. 392--395. Association for Computational Linguistics, Lisbon,
  Portugal (Sep 2015). \doi{10.18653/v1/W15-3049}

\bibitem{10.5555/3455716.3455856}
Raffel, C., Shazeer, N., Roberts, A., Lee, K., Narang, S., Matena, M., Zhou,
  Y., Li, W., Liu, P.J.: Exploring the limits of transfer learning with a
  unified text-to-text transformer. J. Mach. Learn. Res.  \textbf{21}(1) (jan
  2020)

\bibitem{rao-tetreault-2018-dear}
Rao, S., Tetreault, J.: Dear sir or madam, may {I} introduce the {GYAFC}
  dataset: Corpus, benchmarks and metrics for formality style transfer. In:
  Proceedings of the 2018 Conference of the North {A}merican Chapter of the
  Association for Computational Linguistics: Human Language Technologies,
  Volume 1 (Long Papers). pp. 129--140. Association for Computational
  Linguistics, New Orleans, Louisiana (Jun 2018)

\bibitem{rastogi2020towards}
Rastogi, A., Zang, X., Sunkara, S., Gupta, R., Khaitan, P.: Towards scalable
  multi-domain conversational agents: The schema-guided dialogue dataset. In:
  Proceedings of the AAAI Conference on Artificial Intelligence. vol.~34, pp.
  8689--8696 (2020)

\bibitem{reid-zhong-2021-lewis}
Reid, M., Zhong, V.: {LEWIS}: {L}evenshtein editing for unsupervised text style
  transfer. In: Findings of the Association for Computational Linguistics:
  ACL-IJCNLP 2021. pp. 3932--3944. Association for Computational Linguistics,
  Online (Aug 2021)

\bibitem{reif-etal-2022-recipe}
Reif, E., Ippolito, D., Yuan, A., Coenen, A., Callison-Burch, C., Wei, J.: A
  recipe for arbitrary text style transfer with large language models. In:
  Proceedings of the 60th Annual Meeting of the Association for Computational
  Linguistics (Volume 2: Short Papers). pp. 837--848. Association for
  Computational Linguistics, Dublin, Ireland (May 2022)

\bibitem{sellam-etal-2020-bleurt}
Sellam, T., Das, D., Parikh, A.: {BLEURT}: Learning robust metrics for text
  generation. In: Proceedings of the 58th Annual Meeting of the Association for
  Computational Linguistics. pp. 7881--7892. Association for Computational
  Linguistics, Online (Jul 2020)

\bibitem{NIPS2017_2d2c8394}
Shen, T., Lei, T., Barzilay, R., Jaakkola, T.: Style transfer from non-parallel
  text by cross-alignment. In: Guyon, I., Luxburg, U.V., Bengio, S., Wallach,
  H., Fergus, R., Vishwanathan, S., Garnett, R. (eds.) Advances in Neural
  Information Processing Systems. vol.~30. Curran Associates, Inc. (2017)

\bibitem{46201}
Vaswani, A., Shazeer, N., Parmar, N., Uszkoreit, J., Jones, L., Gomez, A.N.,
  Kaiser, L., Polosukhin, I.: Attention is all you need (2017)

\bibitem{warstadt2018neural}
Warstadt, A., Singh, A., Bowman, S.R.: Neural network acceptability judgments.
  arXiv preprint arXiv:1805.12471  (2018)

\bibitem{10.2307/3001968}
Wilcoxon, F.: Individual comparisons by ranking methods. Biometrics Bulletin
  \textbf{1}(6),  80--83 (1945)

\bibitem{ijcai2019-732}
Wu, X., Zhang, T., Zang, L., Han, J., Hu, S.: Mask and infill: Applying masked
  language model for sentiment transfer. In: Proceedings of the Twenty-Eighth
  International Joint Conference on Artificial Intelligence, {IJCAI-19}. pp.
  5271--5277. International Joint Conferences on Artificial Intelligence
  Organization (7 2019)

\bibitem{zhang-etal-2021-dont}
Zhang, M., Tan, L., Fu, Z., Xiong, K., Lin, J., Li, M., Tu, Z.: Don{'}t change
  me! user-controllable selective paraphrase generation. In: Proceedings of the
  16th Conference of the European Chapter of the Association for Computational
  Linguistics: Main Volume. pp. 3522--3527. Association for Computational
  Linguistics, Online (Apr 2021)

\bibitem{zhang-etal-2020-parallel}
Zhang, Y., Ge, T., Sun, X.: Parallel data augmentation for formality style
  transfer. In: Proceedings of the 58th Annual Meeting of the Association for
  Computational Linguistics. pp. 3221--3228. Association for Computational
  Linguistics, Online (Jul 2020)

\bibitem{zhang-etal-2020-pointer}
Zhang, Y., Wang, G., Li, C., Gan, Z., Brockett, C., Dolan, B.: {POINTER}:
  Constrained progressive text generation via insertion-based generative
  pre-training. In: Proceedings of the 2020 Conference on Empirical Methods in
  Natural Language Processing (EMNLP). pp. 8649--8670. Association for
  Computational Linguistics, Online (Nov 2020)

\bibitem{zhuang-etal-2021-robustly}
Zhuang, L., Wayne, L., Ya, S., Jun, Z.: A robustly optimized {BERT}
  pre-training approach with post-training. In: Proceedings of the 20th Chinese
  National Conference on Computational Linguistics. pp. 1218--1227. Chinese
  Information Processing Society of China, Huhhot, China (Aug 2021)

\end{thebibliography}

% \end{thebibliography}

% \appendix 

% \section{Example of models inference}

\end{document}